\documentclass{article}

\usepackage[preprint]{ARLET_2025}
\usepackage[utf8]{inputenc} % allow utf-8 input
\usepackage[T1]{fontenc}    % use 8-bit T1 fonts
\usepackage{hyperref}       % hyperlinks
\usepackage{url}            % simple URL typesetting
\usepackage{booktabs}       % professional-quality tables
\usepackage{amsfonts}       % blackboard math symbols
\usepackage{nicefrac}       % compact symbols for 1/2, etc.
\usepackage{microtype}      % microtypography
\usepackage{xcolor}         % colors

\usepackage{graphicx}
\usepackage{algorithm}
\usepackage{algorithmic}
\usepackage{amsmath}
\usepackage{amssymb}
\usepackage{float}
\usepackage{multirow}
\usepackage{comment}

\title{MAGIC-MASK: Multi-Agent Guided Inter-Agent Collaboration with Mask-Based Explainability for Reinforcement Learning}

\author{
  Maisha Maliha \\
  School of Computer Science \\
  University of Oklahoma \\
  Norman, Oklahoma, USA \\
  \texttt{maisha.maliha-1@ou.edu}
  \And
  Dean Hougen \\
  School of Computer Science \\
  University of Oklahoma \\
  Norman, Oklahoma, USA \\
  \texttt{hougen@ou.edu}
}

\begin{document}

\maketitle

\begin{abstract}
Understanding the decision-making process of Deep Reinforcement Learning agents remains a key challenge for deploying these systems in safety-critical and multi-agent environments. While prior explainability methods like StateMask, have advanced the identification of critical states, they remain limited by computational cost, exploration coverage, and lack of adaptation to multi-agent settings. To overcome these limitations, we propose a mathematically grounded framework, MAGIC-MASK (Multi-Agent Guided Inter-agent Collaboration with Mask-Based Explainability for
Reinforcement Learning), that extends perturbation-based explanation to Multi-Agent Reinforcement Learning. Our method integrates Proximal Policy Optimization, adaptive $\epsilon$-greedy exploration, and lightweight inter-agent collaboration to share masked state information and peer experience. This collaboration enables each agent to perform saliency-guided masking and share reward-based insights with peers, reducing the time required for critical state discovery, improving explanation fidelity, and leading to faster and more robust learning. The core novelty of our approach lies in generalizing explainability from single-agent to multi-agent systems through a unified mathematical formalism built on trajectory perturbation, reward fidelity analysis, and Kullback–Leibler divergence regularization. This framework yields localized, interpretable explanations grounded in probabilistic modeling and multi-agent Markov decision processes. We validate our framework on both single-agent and multi-agent benchmarks, including a multi-agent highway driving environment and Google Research Football, demonstrating that MAGIC-MASK consistently outperforms state-of-the-art baselines in fidelity, learning efficiency, and policy robustness while offering interpretable and transferable explanations.
\end{abstract}

\section{Introduction}

Reinforcement Learning (RL) has shown remarkable success in solving complex sequential decision-making tasks across domains such as robotics, control systems, and games. As RL systems scale from single-agent setups to Multi-Agent Reinforcement Learning (MARL) paradigms, they enable high-impact applications, including unmanned aerial vehicle swarms \cite{liu2022multi, lv2024efficient}, industrial robotics \cite{luo2023joint, qiu2023multiagent}, camera surveillance networks \cite{ci2023proactive}, and autonomous driving \cite{petrillo2018adaptive}. However, despite this progress, Deep Reinforcement Learning (DRL) policies often operate as black boxes, making their decision-making processes difficult to interpret or trust, particularly in multi-agent environments where outcomes depend on intricate cross-agent interactions. Existing explainability techniques, such as credit assignment or value decomposition, offer partial insights, but generally require access to transparent model structures or are limited to in-training diagnostics, making them impractical in black-box MARL settings. Post-hoc explanation methods, including saliency maps \cite{puri2019explain, mccalmon2022caps} and time-step relevance models \cite{huang2018establishing}, have been developed to highlight important input features or critical temporal regions, yet most are designed solely for single-agent systems and fail to capture inter-agent dependencies, stochasticity, and coordination challenges inherent to MARL. Moreover, recent perturbation-based approaches, such as StateMask \cite{cheng2023statemask}, improve critical state identification by selectively randomizing actions and observing performance changes, but they too do not scale effectively to multi-agent scenarios. 

To address these limitations, we propose MAGIC-MASK, a novel explainability framework designed specifically for multi-agent systems. Our method operates under the assumption that only observable trajectories (states, actions, rewards) are accessible, without requiring internal access to policy parameters or value functions. By formalizing the explanation problem within a multi-agent partially observable Markov decision process (POMDP) \cite{kaelbling1998planning}, MAGIC-MASK integrates saliency-guided masking, adaptive $\epsilon$-greedy exploration \cite{garcia2015comprehensive, liu2023maximize}, and lightweight inter-agent collaboration to identify jointly critical states across agents. In our work, saliency-guided masking refers to a perturbation-based approach \cite{cheng2023statemask}, in which the importance of state-action pairs is inferred by systematically modifying actions and measuring the resulting reward variations. Significant drops in reward indicate critical decisions, thereby aligning the method with saliency principles in a performance-grounded manner. Crucially, our framework leverages Proximal
Policy Optimization (PPO) as the backbone, ensuring stable policy updates under both masked and unmasked conditions while preserving agent performance and explanation fidelity. We evaluate MAGIC-MASK across diverse MARL environments, demonstrating its effectiveness in uncovering interpretable decision patterns, improving exploration efficiency, and advancing trust in MARL systems.

\vspace{0.5em}
\noindent \textbf{Our key contributions are:}
\begin{itemize}
    \item We introduce a coordinated multiagent saliency masking framework that identifies jointly critical states across agents by combining local perturbations with inter-agent dependency modeling, enabling explanations that go beyond isolated single-agent perspectives.
    \item We design an adaptive exploration and collaboration protocol where agents share masked state information with peers, reducing redundant exploration and accelerating collective learning in large state-action spaces.
    \item We integrate this framework into a PPO-based training pipeline and provide rigorous empirical validation across diverse multiagent benchmarks, demonstrating superior explanation fidelity, improved critical state discovery, and strong policy performance.
\end{itemize}

\section{Related Work}
\label{sec:related-work}

Explainability in RL has emerged as a critical research area to increase the transparency, trust, and usability of learned policies, especially in high-stakes applications and complex multi-agent environments. Prior research can broadly be categorized into two major branches: in-training explanation methods and post-training black-box explanations.

\textbf{In-training explainable RL methods} embed interpretability directly into the learning process. This includes hierarchical RL approaches \cite{zhang2020generating}, model approximation frameworks \cite{bewley2021tripletree}, and credit assignment strategies such as Shapley value-based methods \cite{li2021shapley}, which distribute responsibility among agents. Notably, COMA \cite{foerster2018counterfactual} connects counterfactual reasoning to credit assignment, helping explain MARL systems. However, these approaches often require access to policy internals, making them unsuitable for post-hoc explanations in black-box settings and frequently produce explanations as secondary byproducts rather than as primary objectives \cite{jacq2022lazy}.

\textbf{Post-training explanation methods} focus on interpreting agent behavior after learning, treating the trained policy as a black box. These approaches operate at two levels. Observation-level techniques highlight salient input features or spatial-temporal regions that most influence agent decisions. These include saliency maps \cite{puri2019explain}, strategy representations \cite{mccalmon2022caps}, or summarization models such as ValueMax \cite{amir2018highlights, huang2018establishing}, which extract meaningful behavioral summaries to communicate agent intent to humans. Step-level methods, on the other hand, identify critical time steps or state sequences pivotal to final outcomes, using tools like Lazy-MDPs \cite{jacq2022lazy}, which selectively model when actions matter, and perturbation-based techniques such as StateMask \cite{cheng2023statemask}, which quantify the importance of state-reward associations. Critical states frameworks \cite{huang2018establishing} have also been developed to build appropriate trust by pinpointing pivotal decision moments. Despite their successes, these methods remain limited in multi-agent contexts where decision-making responsibility must be attributed across interacting agents, not just over time.

\textbf{Counterfactual reasoning and multi-agent attribution} play an increasingly important role in MARL interpretability. While counterfactual methods have been widely explored in supervised learning \cite{goyal2019counterfactual} and adapted for RL at both observation and state levels \cite{louhichi2023shapley, cheng2023statemask}, their extension to multi-agent systems is non-trivial due to coupled dynamics and scalability constraints. Shapley-based explanations provide principled marginal contribution estimates \cite{louhichi2023shapley}, but they scale poorly as the number of agents increases \cite{kumar2020problems}. Recent approaches such as EDGE \cite{guo2021edge} and Lazy-MDPs \cite{jacq2022lazy} begin to address these issues, but they do not deliver fine-grained, agent-level, time-resolved attribution in multi-agent systems, leaving a critical gap. Counterfactual Shapley values have been used to analyze individual agents \cite{chen2025understanding}, but such approaches are post hoc, computationally expensive, and poorly scalable. In contrast, MAGIC-MASK identifies temporally salient states during training and embeds explainability into learning, yielding scalable, policy-aware, collaborative multi-agent explanations.

To address these challenges, our work addresses this underexplored direction by developing a counterfactual, agent-level, time-step-resolved explanation framework specifically developed for MARL environments. By building upon perturbation techniques, adaptive exploration, and inter-agent collaboration, we deliver scalable, high-fidelity multi-agent explanations that advance beyond the current limitations of existing state-of-the-art methods.

\section{Key Technique}
\label{headings}

In this section, we introduce our proposed MAGIC-MASK framework, which extends the original StateMask method \cite{cheng2023statemask} to the MARL setting. Our approach integrates coordinated exploration, inter-agent collaboration, adaptive masking, and mathematically grounded explainability, addressing the unique challenges of multiagent systems where agents interact not only with the environment but also with each other.

\subsection{Overview}
We focus on the problem of identifying critical states \cite{guo2021edge} in multiagent DRL environments, where multiple agents act concurrently, influencing both local outcomes and the collective system behavior \cite{highway-env, machado2018revisiting}. Compared to single-agent systems, multiagent environments introduce additional complexity due to partial observability, cross-agent dependencies, coupled decision-making, and shared resource dynamics, which amplify the interpretability challenge \cite{huh2023multi}. To address these challenges, our framework is designed under the assumption that we have access only to observable trajectories, specifically the sequences of agent-specific states $s_t^i$, actions $a_t^i$, and rewards $r_t^i$, without requiring access to the internal policy parameters $\pi_\theta^i$ or value functions $V^\pi$. We model each agent’s environment as a POMDP \cite{kaelbling1998planning}, where stochastic transitions depend both on the agent’s own actions and the actions of its peers. To ensure scalability, agents interact with their own parallel copies of the environment, enabling efficient and distributed data collection.

We begin by constructing a masking module inspired by the methodology proposed for StateMask \cite{cheng2023statemask}, where critical states are identified by systematically randomizing an agent’s actions and observing the resulting reward perturbations. States that lead to substantial reward changes under randomized actions are marked as critical, while those with negligible impact are considered non-critical. This yields a saliency map over the agent’s observation space. We extend this single-agent explanation approach into a multiagent formulation, introducing MAGIC-MASK, a coordinated explanation system that identifies jointly critical states across agents. To support this, we design a multiagent-aware masking mechanism that captures both individual saliencies and inter-agent dependencies driving global behavior. We address the exploration–exploitation tradeoff \cite{garcia2015comprehensive,liu2023maximize} using adaptive $\epsilon$-greedy exploration, ensuring broad and diverse state-action coverage. Crucially, we incorporate an inter-agent collaboration protocol where all agents actively share their masked state information, allowing them to learn from their peers and accelerate collective learning. This shared knowledge reduces redundant exploration and promotes faster convergence across the system.

\begin{figure*}[t]
    \centering
    \includegraphics[width=\textwidth]{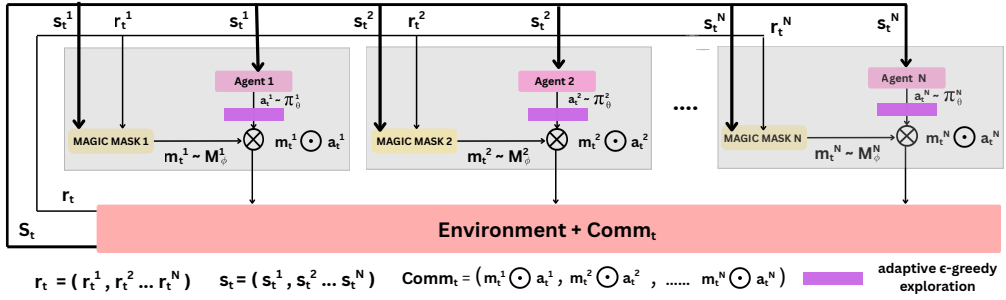} % update the path as needed
    \caption{Overview of the MAGIC-MASK framework.}
    \label{fig:magic_mask_overview}
\end{figure*}

Importantly, we employ PPO as the backbone learning algorithm, leveraging its clipped surrogate objective to enable stable policy updates under both masked and unmasked conditions. This ensures that the optimization process balances performance preservation, policy smoothness, and saliency-guided perturbations in a principled and scalable manner. Finally, we empirically evaluate the impact of masked rollouts on multiagent performance, providing a general framework for explaining the decision dynamics and behaviors of RL systems. An overview of the MAGIC-MASK framework is illustrated in Figure~\ref{fig:magic_mask_overview}, where each agent applies masking to its local state, perturbs actions based on mask output, and interacts with the environment. The collective masked states are shared via the $Comm_t$ signal. Adaptive $\epsilon$-greedy exploration promotes diverse sampling across agents. Here, $s_t$ denotes the joint state and $r_t$ the joint reward at time step $t$, both aggregated across agents.

\subsection{Mathematical Formulation}

We formalize the multiagent explainability problem using an extended explanation Markov Decision Process (MDP), designed to capture the dynamics of critical state identification under masking. Let there be $N$ agents, where each agent $i$ interacts with its local state $s_t^i$ and selects a local action $a_t^i$ at time $t$. Together, these form the global environment state $S_t = (s_t^1, s_t^2, \dots, s_t^N)$ and the joint action vector $A_t = (a_t^1, a_t^2, \dots, a_t^N)$. Each agent’s behavior is governed by a policy $\pi_\theta^i(a_t^i | s_t^i)$, parameterized by neural network weights $\theta^i$, which determines its action distribution conditioned on local observations. To enable explainability, we equip each agent with a state mask network $M_\phi^i(s_t^i)$, parameterized by $\phi^i$, which outputs a soft masking probability $m_t^i \in [0,1]$. This mask determines whether the agent’s action at a given step is drawn from its learned policy or replaced with a randomized baseline, providing a controlled mechanism for probing the importance of specific states. The masked policy is formally defined as a conditional action selection rule: when $m_t^i > \tau$, the agent follows its policy’s predicted action $a_t^i \sim \pi_\theta^i(a_t^i | s_t^i)$; when $m_t^i \leq \tau$, the agent switches to a uniformly random action sampled from the action space $\mathcal{A}$. Here, $\tau$ is a tunable randomization threshold (set to 0.5, a commonly adopted midpoint that balances perturbation and stability, and has shown robust performance across all evaluated environments). By selectively randomizing actions at targeted states, the framework reveals which time steps are pivotal for downstream performance. The optimization goal is to learn $M_\phi^i$ such that the masked agent’s cumulative reward $R_i^{\pi, M}$ remains as close as possible to the original unmasked reward $R_i^\pi$. Formally, the objective is to minimize the expected reward deviation:
\begin{equation}
%\[
\min_{\phi^i} \; \mathbb{E} \left[ \left| R_i^\pi - R_i^{\pi, M} \right| \right].
%\]
\end{equation}
However, directly optimizing over discrete masking decisions is infeasible due to non-differentiability. To address this, we introduce a continuous surrogate loss function:
\begin{equation}
%\[
\mathcal{L}(\phi^i) = \mathrm{MSE}\left( \mathbb{E}[m_t^i], \tau \right),
\label{eq:mask_loss}
%\]
\end{equation}
which encourages the average mask outputs across time steps to align with the threshold $\tau$, thereby balancing determinism and randomization. This formulation naturally extends the single-agent StateMask idea into a multiagent context, where the challenge is no longer just local state masking
but the coordinated identification of jointly critical states across agents.
Embedding masking behavior into the Explanation MDP lays the foundation for integrating optimization, communication, and explainability into a unified mathematical model. This is essential for MAGIC-MASK’s success, as it supports both localized, agent-specific explanations and collective multiagent behavior analysis, ultimately enabling transparent and trustworthy reinforcement learning systems.

\subsection{Technical Implementation Details}

\paragraph{Agent Policy Optimization.} We adopt PPO as the backbone for both policy learning and mask network training in the MAGIC-MASK framework. PPO is selected due to its stable on-policy learning dynamics, particularly its use of the clipped surrogate objective:
\begin{equation}
%\[
\mathcal{L}^{PPO}(\theta^i) = \mathbb{E}_t \left[ \min \left( r_t^i(\theta^i) \hat{A}_t^i, \; \mathrm{clip}(r_t^i(\theta^i), 1 - \epsilon, 1 + \epsilon) \hat{A}_t^i \right) \right],
%\]
\end{equation}
where
\begin{equation*}
    r_t^i(\theta^i) = \frac{\pi_\theta^i(a_t^i | s_t^i)}{\pi_{\theta_\text{old}}^i(a_t^i | s_t^i)}
\end{equation*} is the probability ratio between the current and old policies, and $\hat{A}_t^i$ is the estimated advantage. The advantage, a key signal in PPO, quantifies how much better or worse an action performed compared to the expected value under the baseline, often computed using generalized advantage estimation (GAE) that balances bias and variance. This clipped surrogate loss prevents large policy updates, stabilizing learning, which is crucial when multiple agents are trained concurrently.

\paragraph{Mask Network Optimization and Adaptive Exploration.}  
Each agent’s \textit{mask network} \( M_\phi^i \) is responsible for generating binary decisions \( m_t^i \in \{0,1\} \) that determine whether to perturb the action at a given timestep. Since direct optimization over binary outputs is non-differentiable, we use a surrogate mean-squared error (MSE) loss, as shown in Equation~\ref{eq:mask_loss}. This formulation encourages the network to selectively apply perturbations only at non-critical states, allowing the agent to preserve behavior in important states and randomize elsewhere. As a result, the mask network effectively identifies temporally salient decision points without destabilizing the learning process.

%where \( \tau \) is a predefined sparsity threshold. 
To further promote diverse state-action coverage, we incorporate adaptive \( \epsilon \)-greedy exploration using
\begin{equation}
\epsilon_t^i = \epsilon_0 \cdot \exp(-\lambda t),
\end{equation}
where \(\epsilon_0\) is the initial value at time \(t=0\). At each timestep, with probability \( \epsilon_t^i \), the agent selects a random action \( a_t^i \sim \mathrm{Uniform}(\mathcal{A}) \); otherwise, it chooses the action according to its learned policy \( a_t^i \sim \pi_\theta^i(s_t^i) \). The value of \( \epsilon_t^i \) is dynamically adjusted based on the progress of the training or empirical uncertainty to balance exploration and exploitation. This stochastic exploration mechanism complements the saliency-guided masking strategy (in which masking is guided by reward-based perturbation outcomes) and facilitates efficient discovery of critical states throughout training.

\paragraph{Inter-Agent Collaboration.} A distinctive feature of MAGIC-MASK is the inter-agent collaboration protocol, centered around the shared communication signal $\mathrm{Comm}_t$. After each parallel rollout phase, each agent collects its set of masked (perturbed) states where $m_t^i \leq \tau$, assembling a local masked state set $\mathrm{Comm}_t^i$. These local sets are then broadcast to all peers, where they are aggregated:
%\[
\begin{equation}
\mathrm{Comm}_t = \bigcup_i \mathrm{Comm}_t^i.
%\]
\end{equation}
This global masked state set represents the collective knowledge about which regions of the state space have been explored and perturbed, revealing where saliency mapping has already been sufficiently covered. 
The $\mathrm{Comm}_t$ signal plays a critical role in shaping subsequent learning phases. Rather than being fed directly as input to neural networks, it is used to inform exploration policies and mask network adjustments. Specifically, each agent leverages the aggregated $\mathrm{Comm}_t$ to prioritize underexplored states in the next rollout, reducing redundant masking of already-known critical regions. This coordination accelerates convergence, improves exploration efficiency, and promotes knowledge sharing among agents, aligning the entire system toward discovering jointly critical states that impact both local and global performance. By tightly integrating PPO’s advantage-guided updates, surrogate loss optimization for masking, and inter-agent communication through $\mathrm{Comm}_t$, the MAGIC-MASK framework establishes a principled pipeline for multiagent interpretability. Each mathematical component serves a distinct role: PPO ensures stable policy learning, the mask loss identifies salient temporal regions, and the collaboration protocol weaves together agents’ discoveries into a unified critical state map. This collective mechanism ultimately delivers high-fidelity explanations of policy behavior across complex MARL environments.

\begin{algorithm}[t]
\caption{MAGIC-MASK Framework}
\fbox{\parbox{\columnwidth}{
\textbf{Input:} $N$ agents with policies $\pi_\theta^i$, mask networks $M_\phi^i$, environments $E^i$; randomization threshold $\tau$; exploration rate $\epsilon$; learning rates $\eta_\theta$, $\eta_\phi$; maximum iterations $T_{\text{max}}$ \\
\textbf{Output:} Optimized policy parameters $\theta^i$ and mask parameters $\phi^i$ \\
\textbf{Initialization:} Initialize policies $\pi_\theta^i$ and masks $M_\phi^i$ randomly \\
\textbf{for} iteration $t = 1$ to $T_{\text{max}}$ \textbf{do} \\
\hspace*{0.5em} \textbf{Parallel rollout phase:} \\
\hspace*{1em} For each agent $i$: \\
\hspace*{2em} Observe local state $s_t^i$ \\
\hspace*{2em} Compute mask probability $m_t^i = M_\phi^i(s_t^i)$ \\
\hspace*{2em} If $m_t^i > \tau$, take $a_t^i \sim \pi_\theta^i(a_t^i | s_t^i)$; else $a_t^i \sim \mathrm{Uniform}(\mathcal{A}^i)$ \\
\hspace*{2em} Apply adaptive $\epsilon$-greedy exploration \\
\hspace*{2em} Execute $a_t^i$ in $E^i$, store $(s_t^i, a_t^i, r_t^i, s_{t+1}^i)$ in rollout buffer \\
\hspace*{0.5em} \textbf{Policy optimization phase (PPO):} \\
\hspace*{1em} For each agent $i$: \\
\hspace*{2em} Compute advantage estimates $\hat{A}_t^i$ \\
\hspace*{2em} Compute PPO clipped surrogate loss: \\
\hspace*{3em} $\mathcal{L}^{PPO}(\theta^i) = \mathbb{E}_t \left[ \min \big( r_t^i(\theta^i) \hat{A}_t^i, \mathrm{clip}(r_t^i(\theta^i), 1 - \epsilon, 1 + \epsilon) \hat{A}_t^i \big) \right]$ \\
\hspace*{2em} Update $\theta^i \gets \theta^i - \eta_\theta \nabla_{\theta^i} \mathcal{L}^{PPO}(\theta^i)$ \\
%\hspace*{2em} Record $D_{\mathrm{KL}}(\pi_{\text{old}}\|\pi_{\text{new}})$ on the rollout batch \\
\hspace*{0.5em} \textbf{Mask optimization phase:} \\
\hspace*{1em} For each agent $i$: \\
\hspace*{2em} Compute surrogate mask loss: \\
\hspace*{3em} $\mathcal{L}(\phi^i) = \mathrm{MSE}(\mathbb{E}[m_t^i], \tau)$ \\
\hspace*{2em} add reward preservation and Kullback–Leibler (KL) divergence terms  \\
\hspace*{2em} Update $\phi^i \gets \phi^i - \eta_\phi \nabla_{\phi^i} \mathcal{L}(\phi^i)$ \\
\hspace*{0.5em} \textbf{Inter-agent collaboration phase:} \\
\hspace*{1em} For each agent $i$: \\
\hspace*{2em} Broadcast masked critical states $\{ s_t^i | m_t^i \leq \tau \}$ \\
\hspace*{2em} Update shared exploration weights or priority buffers across agents \\
\textbf{end for} \\
\textbf{Evaluation:} Periodically assess reward preservation, KL divergence, and explanation fidelity over held-out validation episodes \\
\textbf{Termination:} Stop when $t = T_{\text{max}}$ or performance convergence is reached
}}
\end{algorithm}

%\newpage
\section{Evaluation}

To evaluate the effectiveness and generalizability of MAGIC-MASK, we design experiments across a range of diverse reinforcement learning environments and benchmark settings.

\subsection{Evaluation Environments}

To comprehensively test MAGIC-MASK, we select five environments that vary in state space structure, observability, and coordination complexity: Connect 4 \cite{allis1988knowledge}, Doudizhu \cite{zha2021douzero}, Pong \cite{machado2018revisiting}, Multi-Agent Highway \cite{highway-env} and Google Research Football \cite{kurach2020google}. These tasks represent symbolic board games, multi-agent card games, visual-based control, and real-world-inspired autonomous driving scenarios, respectively. This variety enables a robust assessment of MAGIC-MASK’s adaptability to discrete and continuous settings, with varying agent dynamics and partial observability. All agents are trained using PPO, with MAGIC-MASK agents incorporating a saliency-aware masking module and an inter-agent collaboration mechanism. Adaptive $\epsilon$-greedy exploration is used to promote exploration efficiency. Key metrics for evaluation include final average reward, KL divergence (policy shift post-masking), inter-agent fidelity, and reward drop after trajectory perturbations. For information on environments, see Appendix\ref{app8}.

\begin{comment}
\paragraph{Multi-Agent Highway Environment.}
The \textit{Multi-Agent Highway} environment is implemented using the \texttt{multiagent\_highway} \cite{highway-env} project, simulates autonomous vehicle coordination in dynamic traffic settings. Each agent learns to manage lane switching, acceleration, and collision avoidance in a shared road environment. With \textsc{MAGIC-MASK}, agents identify risk-sensitive states via action perturbation and communicate saliency findings with each other. This setup highlights \textsc{MAGIC-MASK}’s effectiveness in high-dimensional, partially observable MARL environments where safety and coordination are crucial.
\end{comment}

\paragraph{Baseline Selection.}
We benchmark MAGIC-MASK against four state-of-the-art explainability approaches: StateMask \cite{cheng2023statemask}, LazyMDP \cite{jacq2022lazy}, EDGE \cite{guo2021edge}, and ValueMax \cite{amir2018highlights, huang2018establishing}. StateMask serves as a strong perturbation-based saliency baseline in single-agent settings, while LazyMDP focuses on explainability through sparse action selection. EDGE emphasizes identifying critical states by measuring reward sensitivity, and ValueMax offers a smooth saliency approximation using soft-value propagation. These baselines represent leading methods in reinforcement learning interpretability, making them suitable benchmarks for assessing MAGIC-MASK’s performance. All baseline methods are implemented following their original designs to ensure consistent comparisons.

\subsection{Experiment Design}

\begin{figure}[H]
    \centering
    \makebox[\textwidth]{%
        \includegraphics[width=1\textwidth]{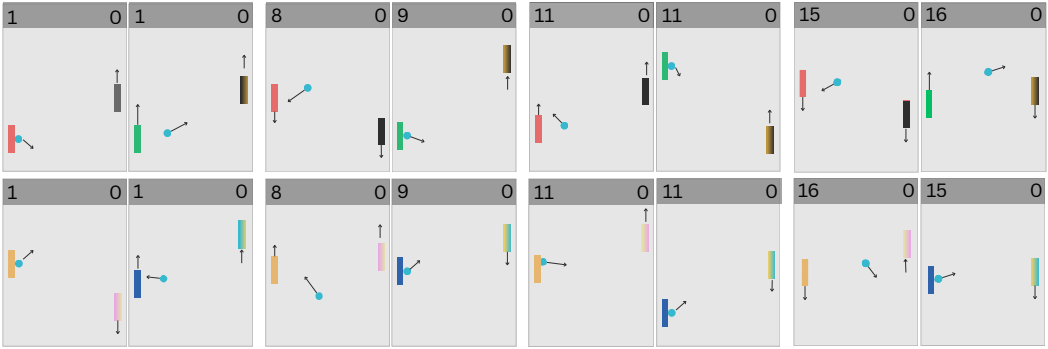}
    }
    \caption{Visualization of MAGIC-MASK in the multi-agent Pong environment.}
    \label{fig:pong-visualization}
\end{figure}

For each of the five environments, we train MAGIC-MASK and baseline agents for 10 million timesteps under identical hyperparameter and architecture settings (for further details, see Appendix~\ref{app3}). We measure not only raw reward performance but also robustness through post-training policy perturbations and stability using KL divergence metrics. Fidelity between agents is assessed to evaluate how consistently agents align under each approach. The experimental design is structured to isolate the effects of MAGIC-MASK’s multi-agent saliency and collaboration mechanisms. In Figure~\ref{fig:pong-visualization}, each panel depicts a different agent playing in its own environment copy, capturing unique saliency-triggered behavior based on randomized perturbations. The left block of panels shows agents trained with our MAGIC-MASK-based DRL framework, while the right block displays agents trained using standard reinforcement learning. Each row corresponds to one of four identical environments observed across different episodes, enabling a fair visual comparison. By observing reward changes in these localized rollouts, MAGIC-MASK agents identify critical states more effectively. The collected saliency information is then shared across agents to guide exploration and enhance explanation fidelity through peer experience.

\begin{comment}
This collaborative process reduces redundant exploration and accelerates convergence, contributing to higher reward and better interpretability compared to standard RL agents.

In Figure ~\ref{fig:pong-visualization}, each panel depicts a different agent playing in its own environment copy, capturing unique saliency-triggered behavior based on randomized perturbations. By observing reward changes in these localized rollouts, agents identify critical states. The collected saliency information is then shared across agents to guide exploration and enhance explanation fidelity through peer experience. This collaborative process reduces redundant exploration and accelerates convergence.
\end{comment}

\begin{figure}[ht]
    \centering
    \includegraphics[width=\textwidth]{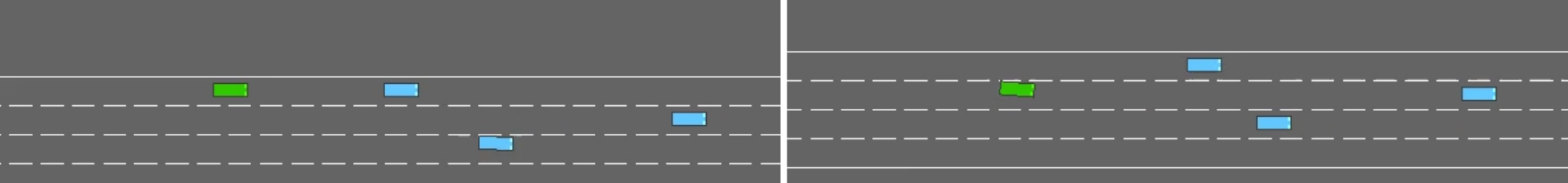} % Update path as needed
    \caption{Illustration of MAGIC-MASK applied to the multi-agent highway environment.} 
    \label{fig:highway-visualization}
\end{figure}

In Figure~\ref{fig:highway-visualization}, the image shows an agent’s observation with highlighted regions corresponding to saliency-based masking. Figure 3, compares decision-making behaviors between a standard DRL agent (left) and our proposed MAGIC-MASK-enhanced DRL agent (right). In the left panel, the green vehicle, controlled by a baseline DRL agent, selects a lane with minimal inter-vehicle distance. This behavior illustrates the model’s limited situational awareness, as the agent’s choice leads to a highly accident-prone state by merging into dense traffic without accounting for safety margins. By contrast, in the right panel, the green vehicle trained with MAGIC-MASK, exhibits more strategic behavior by selecting a lane with maximal inter-vehicle distance. This outcome reflects how saliency-guided communication between agents enables the model to recognize risk-sensitive states and prioritize collision-avoidant trajectories. Consequently, MAGIC-MASK facilitates improved coordination and safety, demonstrating its effectiveness in partially observable multi-agent driving environments. %Blue regions indicate critical states identified by the agent during its own trajectory, where perturbing the action led to a significant reward drop. Green regions represent critical states shared by peer agents through inter-agent collaboration. %
These shared saliency signals help agents anticipate and respond to risky situations they have not directly experienced, such as close interactions with nearby vehicles. This collaborative exchange of masked states reduces redundant exploration and contributes to faster convergence and improved explanation fidelity.

\subsection{Experimental Results}

\begin{comment}
We present a quantitative comparison of \textsc{MAGIC-MASK} against four state-of-the-art explainability methods: StateMask, LazyMDP, EDGE, and ValueMax across four diverse environments: Connect Four, Pong, Doudizhu, and Multiagent Highway. Each model was trained using PPO for 10 million timesteps under identical hyperparameters and evaluation conditions to ensure fairness.
\end{comment}

\textbf{Final Average Reward.} This metric reflects overall policy performance averaged over evaluation episodes. MAGIC-MASK achieves the highest average reward across all environments. In Pong, MAGIC-MASK reaches 22.5, significantly outperforming StateMask (19.8) and ValueMax (21.0), demonstrating improved policy optimization through saliency-driven collaboration. In Doudizhu and Multiagent Highway, MAGIC-MASK provides a 14–15\% gain over Statemask, suggesting better coordination and critical state generalization. In Connect 4, the reward improvement is 40.2, exceeding StateMask’s 37.6 gain.

\begin{comment}
\textbf{Final Average Reward.} This metric reflects the overall policy performance averaged over evaluation episodes. \textsc{MAGIC-MASK} achieves the highest average reward across all environments. In Pong, our method reaches 22.5, significantly outperforming StateMask (19.8) and ValueMax (21.0), demonstrating improved policy optimization through saliency-driven collaboration. In Doudizhu and Multiagent Highway, \textsc{MAGIC-MASK} provides a 14--15\% gain over baselines, suggesting better coordination and critical state generalization. For Connect Four, \textsc{MAGIC-MASK} shows a 40.2\% reward improvement, exceeding StateMask’s 37.6\% gain.
\end{comment}

\textbf{KL Divergence.} KL divergence measures the distributional shift between the original policy and the perturbed one. Lower values indicate greater policy stability under explanation-based interventions. MAGIC-MASK achieves the lowest divergence across all tasks (ranging from 0.08 to 0.09), consistently improving over StateMask and the other baseline models.

\begin{table}[h!]
\centering
\resizebox{\textwidth}{!}{%
\begin{tabular}{ccccccc}
\toprule
\textbf{Metric} & \textbf{Environment} & \textbf{MAGIC-MASK} & \textbf{StateMask} & \textbf{LazyMDP} & \textbf{EDGE} & \textbf{ValueMax} \\ \midrule

\multirow{4}{*}{\parbox{0.8in}{\centering Final Average Reward $\uparrow$}} 
& Connect 4 & $\mathbf{40.2}$ & 37.6 & 36.2 & 24.8 & 35.8 \\ %\cmidrule{2-7}
& Pong & $\mathbf{22.5}$ & 19.8 & 18 & 19 & 21 \\ %\cmidrule{2-7}
& Doudizhu & $\mathbf{14}$ & 11 & 10 & 12 & 9 \\ %\cmidrule{2-7}
& Multiagent Hwy & $\mathbf{15}$ & 13 & 8 & 12 & 10 \\ \midrule

\multirow{4}{*}{\parbox{0.8in}{\centering KL Divergence $\downarrow$}} 
& Connect 4 & $\mathbf{0.08}$ & 0.12 & 0.10 & 0.13 & 0.11 \\ %\cmidrule{2-7}
& Pong & $\mathbf{0.08}$ & 0.12 & 0.10 & 0.13 & 0.11 \\ %\cmidrule{2-7}
& Doudizhu & $\mathbf{0.09}$ & 0.12 & 0.10 & 0.13 & 0.11 \\ %\cmidrule{2-7}
& Multiagent Hwy & $\mathbf{0.08}$ & 0.12 & 0.10 & 0.13 & 0.11 \\ \midrule

\multirow{4}{*}{\parbox{0.8in}{\centering Fidelity Between Agents $\uparrow$}} 
& Connect 4 & $\mathbf{0.91}$ & 0.78 & 0.65 & 0.63 & 0.69 \\ %\cmidrule{2-7}
& Pong & $\mathbf{0.92}$ & 0.78 & 0.85 & 0.80 & 0.87 \\ %\cmidrule{2-7}
& Doudizhu & $\mathbf{0.91}$ & 0.78 & 0.85 & 0.80 & 0.87 \\ %\cmidrule{2-7}
& Multiagent Hwy & $\mathbf{0.92}$ & 0.78 & 0.85 & 0.80 & 0.87 \\ \midrule

\multirow{4}{*}{\parbox{0.8in}{\centering Reward Drop\\After Perturbation $\downarrow$}} 
& Connect 4 & $\mathbf{-16.2\%}$ & -11.6\% & -14.9\% & -12.7\% & -10.3\% \\ %\cmidrule{2-7}
& Pong & $\mathbf{-15.1\%}$ & -12.4\% & -11.2\% & -13.1\% & -10.5\% \\ %\cmidrule{2-7}
& Doudizhu & $\mathbf{-15.7\%}$ & -11.9\% & -14.7\% & -13.4\% & -10.1\% \\ %\cmidrule{2-7}
& Multiagent Hwy & $\mathbf{-15.0\%}$ & -12.1\% & -11.8\% & -13.2\% & -10.2\% \\ \bottomrule

\end{tabular}%
}
\caption{Comparative performance of MAGIC-MASK \& baseline methods across four diverse RL environments. All reward drop values are averaged over three runs; standard deviation is $\leq 0.3$ in all cases.}
\end{table}

The evaluation results demonstrate that MAGIC-MASK consistently outperforms baselines across key metrics, achieving higher final rewards, improved policy stability (lower KL divergence), stronger inter-agent fidelity, and greater robustness against perturbations. These findings highlight MAGIC-MASK as a scalable and generalizable explainability framework for MARL, advancing both interpretability and effectiveness of collaborative learning systems.

\textbf{Fidelity Between Agents.} Fidelity captures the consistency of explanations across agents by quantifying how similarly they behave under shared critical state masking. Higher scores reflect stronger inter-agent alignment and coherent explanation behavior. MAGIC-MASK maintains high fidelity in all environments. For example, our model reaches 0.92 in Pong and Multiagent Highway compared to 0.78 by StateMask and 0.80 by EDGE. In Connect 4, MAGIC-MASK scores 0.91, outperforming LazyMDP’s 0.65 and EDGE’s 0.63.

\textbf{Reward Drop After Perturbation.}  
To evaluate which method most effectively identifies critical states, we adopt a perturbation-based strategy: after each method identifies its set of critical time steps, we randomly perturb the agent’s actions at those steps and measure the resulting drop in cumulative reward. In this context, a larger reward drop indicates that the perturbed states had a greater impact on the policy performance, implying that those states were indeed critical or salient. Conversely, a smaller drop suggests the perturbed states were less consequential, weakening the quality of the explanation. MAGIC-MASK shows the smallest degradation across all tasks. In Pong and Highway, the reward drops by 15\%, while StateMask experiences a 12\% drop and EDGE reaches 13\%. For further details on the components, see Appendices\ref{app4}–\ref{app7}.

\begin{comment}
We extend our evaluation to the GRF 3-vs-1 with keeper task \cite{kurach2020google}, which demands continuous inter-agent coordination. We compare MAGIC-MASK against the counterfactual Shapley value method \cite{chen2025understanding}. On GRF, MAGIC-MASK attains a mean reward drop after perturbation of $-3.25\,(\pm 0.03)$, versus $-2.00\,(\pm 0.03)$ for the baseline, averaged over three independent runs of 10M timesteps. Beyond the aggregate scores, MAGIC-MASK exhibits lower variance, stronger coordination, and smaller perturbation-induced reward drops, demonstrating its effectiveness for robust cooperative policy learning in realistic multi-agent environments.
\end{comment}

We extend our evaluation to the Google Research Football (GRF) 3vs1 with keeper task \cite{kurach2020google}, which requires continuous coordination between agents. In this GRF setting, MAGIC-MASK attains a mean reward drop after perturbation of $-3.25\,(\pm 0.03)$, compared with $-2.00\,(\pm 0.03)$ for the baseline \cite{chen2025understanding}, averaged over three independent runs of 10M timesteps. Beyond the aggregate scores, MAGIC-MASK exhibits lower variance, stronger coordination, and smaller perturbation-induced reward drops, demonstrating its effectiveness for robust cooperative policy learning in realistic multi-agent environments. We further analyze the sensitivity of $\tau$ and the contribution of each component in Appendices~\ref{app1} and \ref{app2}, respectively.

\section{Utility of Explanation in Multiagent Systems}

In safety-critical environments such as autonomous driving, explainability is essential for deployment, verification, and trust. In MARL, understanding and attributing agent behavior is even more important due to interactions between multiple decision-makers in a shared environment. MAGIC-MASK provides a structured way to interpret these behaviors in terms of state, action, and reward.

To better illustrate the utility of our explanation mechanism, we consider a multi-agent autonomous driving scenario, as visualized in Figure~\ref{fig:highway-visualization}. For example, when agent \textsc{0} encounters a pedestrian crossing and does not stop, it leads to severe consequences and a strong negative reward due to potential harm. Likewise, agent \textsc{1} collides with a stone in its path, resulting in a vehicle crash and another significant penalty. In contrast, agent \textsc{2} ignores a rabbit on the road and drives through; while ethically questionable, this situation does not cause a crash and results in only a minor reward reduction. Each of these experiences reflects a distinct level of criticality. Through MAGIC-MASK, these agents share their saliency insights, inferred through reward-based perturbations,  so that agent \textsc{1} or \textsc{2}, who has never encountered a pedestrian, can still learn the importance of braking in such situations. This peer-to-peer propagation of masked rollouts leads to faster convergence and helps all agents internalize high-risk scenarios without individually experiencing them.

Rather than treating these episodes as isolated events, MAGIC-MASK facilitates shared learning through randomized action perturbations. By identifying states that lead to significant reward changes, agents uncover critical decision points. These findings are shared across the agent population, reducing redundant exploration and accelerating convergence. This type of explanation offers two primary benefits. First, it links outcomes to specific decisions, enabling detailed insight into both failures and successes. Second, it distributes learning across agents, improving sample efficiency. Compared to methods that either focus on single-agent saliency or neglect inter-agent coordination, MAGIC-MASK provides a unified and scalable framework for interpretable MARL. In domains like Pong and Doudizhu, it highlights key decision frames that influence outcomes, while shared experiences reduce the number of episodes required for convergence. This makes MAGIC-MASK, a general purpose interpretability tool, well suited for environments demanding transparency, efficiency, and coordination.

%Finally, by tying explanations directly to reward dynamics and enabling temporal saliency analysis through masking, \textsc{MAGIC-MASK} generates explanations that are both interpretable and actionable. 

\section{Conclusions}

We propose MAGIC-MASK, a scalable framework for explainability in MARL, that identifies critical decisions via perturbation-based masking rather than model internals, producing reward-driven saliency maps shared among agents to support coordinated learning. By integrating inter-agent communication, adaptive $\epsilon$-greedy exploration, and KL-regularized PPO training, it addresses partial observability and exploration costs that limit single-agent methods like StateMask. Across diverse environments, MAGIC-MASK improves explanation fidelity, convergence, and policy stability, with future extensions aimed at generating human-readable rationales and supporting hierarchical and cooperative multi-agent settings to enhance transparency and trust in real-world RL.

\newpage

\bibliographystyle{unsrt}  
\bibliography{ARLET_2025}

\begin{thebibliography}{99}

\bibitem{liu2022multi}
Da Liu, Liqian Dou, Ruilong Zhang, Xiuyun Zhang, and Qun Zong.
Multi-agent reinforcement learning-based coordinated dynamic task allocation for heterogenous UAVs.
\textit{IEEE Transactions on Vehicular Technology}, 72(4):4372--4383, 2022.

\bibitem{lv2024efficient}
Zefang Lv, Liang Xiao, Yousong Du, Yunjun Zhu, Shuai Han, and Yong-Jin Liu.
Efficient communications in multi-agent reinforcement learning for mobile applications.
\textit{IEEE Transactions on Wireless Communications}, 2024.

\bibitem{luo2023joint}
Ruyu Luo, Wanli Ni, Hui Tian, Julian Cheng, and Kwang-Cheng Chen.
Joint trajectory and radio resource optimization for autonomous mobile robots exploiting multi-agent reinforcement learning.
\textit{IEEE Transactions on Communications}, 71(9):5244--5258, 2023.

\bibitem{qiu2023multiagent}
Changlin Qiu, Zhengxing Wu, Jian Wang, Min Tan, and Junzhi Yu.
Multiagent-reinforcement-learning-based stable path tracking control for a bionic robotic fish with reaction wheel.
\textit{IEEE Transactions on Industrial Electronics}, 70(12):12670--12679, 2023.

\bibitem{ci2023proactive}
Hai Ci, Mickel Liu, Xuehai Pan, Fangwei Zhong, and Yizhou Wang.
Proactive multi-camera collaboration for 3D human pose estimation.
\textit{arXiv preprint arXiv:2303.03767}, 2023.

\bibitem{petrillo2018adaptive}
Alberto Petrillo, Alessandro Salvi, Stefania Santini, and Antonio Saverio Valente.
Adaptive multi-agents synchronization for collaborative driving of autonomous vehicles with multiple communication delays.
\textit{Transportation Research Part C: Emerging Technologies}, 86:372--392, 2018.

\bibitem{puri2019explain}
Nikaash Puri, Sukriti Verma, Piyush Gupta, Dhruv Kayastha, Shripad Deshmukh,
Balaji Krishnamurthy, and Sameer Singh.
Explain your move: Understanding agent actions using specific and relevant feature attribution.
\textit{arXiv preprint arXiv:1912.12191}, 2019.


\bibitem{mccalmon2022caps}
Joe McCalmon, Thai Le, Sarra Alqahtani, and Dongwon Lee.
Caps: Comprehensible abstract policy summaries for explaining reinforcement learning agents.
In \textit{Proceedings of the International Conference on Autonomous Agents and Multiagent Systems (AAMAS)}, 2022.

\bibitem{huang2018establishing}
Sandy H. Huang, Kush Bhatia, Pieter Abbeel, and Anca D. Dragan.
Establishing appropriate trust via critical states.
In \textit{2018 IEEE/RSJ International Conference on Intelligent Robots and Systems (IROS)}, pages 3929--3936. IEEE, 2018.


\bibitem{cheng2023statemask}
Zelei Cheng, Xian Wu, Jiahao Yu, Wenhai Sun, Wenbo Guo, and Xinyu Xing.
Statemask: Explaining deep reinforcement learning through state mask.
\textit{Advances in Neural Information Processing Systems}, 36:62457--62487, 2023.


\bibitem{kaelbling1998planning}
Leslie Pack Kaelbling, Michael L. Littman, and Anthony R. Cassandra.
Planning and acting in partially observable stochastic domains.
\textit{Artificial Intelligence}, 101(1--2):99--134, 1998. Elsevier.

\bibitem{garcia2015comprehensive}
Javier García and Fernando Fernández.
A comprehensive survey on safe reinforcement learning.
\textit{Journal of Machine Learning Research}, 16(1):1437--1480, 2015.


\bibitem{liu2023maximize}
Zhihan Liu, Miao Lu, Wei Xiong, Han Zhong, Hao Hu, Shenao Zhang,
Sirui Zheng, Zhuoran Yang, and Zhaoran Wang.
Maximize to explore: One objective function fusing estimation, planning, and exploration.
\textit{Advances in Neural Information Processing Systems}, 36:22151--22165, 2023.


\bibitem{zhang2020generating}
Tianren Zhang, Shangqi Guo, Tian Tan, Xiaolin Hu, and Feng Chen.
Generating adjacency-constrained subgoals in hierarchical reinforcement learning.
\textit{Advances in Neural Information Processing Systems}, 33:21579--21590, 2020.


\bibitem{bewley2021tripletree}
Tom Bewley and Jonathan Lawry.
Tripletree: A versatile interpretable representation of black box agents and their environments.
In \textit{Proceedings of the AAAI Conference on Artificial Intelligence}, 35(13):11415--11422, 2021.


\bibitem{li2021shapley}
Jiahui Li, Kun Kuang, Baoxiang Wang, Furui Liu, Long Chen, Fei Wu, and Jun Xiao.
Shapley counterfactual credits for multi-agent reinforcement learning.
In \textit{Proceedings of the 27th ACM SIGKDD Conference on Knowledge Discovery \& Data Mining}, pages 934--942, 2021.


\bibitem{foerster2018counterfactual}
Jakob Foerster, Gregory Farquhar, Triantafyllos Afouras, Nantas Nardelli, and Shimon Whiteson.
Counterfactual multi-agent policy gradients.
In \textit{Proceedings of the AAAI Conference on Artificial Intelligence}, 32(1), 2018.

\bibitem{jacq2022lazy}
Alexis Jacq, Johan Ferret, Olivier Pietquin, and Matthieu Geist.
Lazy-MDPs: Towards interpretable RL by learning when to act.
In \textit{Proceedings of the International Conference on Autonomous Agents and Multiagent Systems (AAMAS)}, pages 669--677, 2022.


\bibitem{amir2018highlights}
Dan Amir and Ofra Amir.
Highlights: Summarizing agent behavior to people.
In \textit{Proceedings of the 17th International Conference on Autonomous Agents and Multiagent Systems}, pages 1168--1176, 2018.



\bibitem{goyal2019counterfactual}
Yash Goyal, Ziyan Wu, Jan Ernst, Dhruv Batra, Devi Parikh, and Stefan Lee.
Counterfactual visual explanations.
In \textit{Proceedings of the International Conference on Machine Learning (ICML)}, pages 2376--2384. PMLR, 2019.


\bibitem{louhichi2023shapley}
Mouad Louhichi, Redwane Nesmaoui, Marwan Mbarek, and Mohamed Lazaar.
Shapley values for explaining the black box nature of machine learning model clustering.
\textit{Procedia Computer Science}, 220:806--811, 2023. Elsevier.

\bibitem{kumar2020problems}
I. Elizabeth Kumar, Suresh Venkatasubramanian, Carlos Scheidegger, and Sorelle Friedler.
Problems with Shapley-value-based explanations as feature importance measures.
In \textit{Proceedings of the International Conference on Machine Learning (ICML)}, pages 5491--5500. PMLR, 2020.

\bibitem{chen2025understanding}
Jianming Chen, Yawen Wang, Junjie Wang, Xiaofei Xie, Jun Hu, Qing Wang, and Fanjiang Xu.
Understanding individual agent importance in multi-agent system via counterfactual reasoning.
In \textit{Proceedings of the AAAI Conference on Artificial Intelligence}, 39(15):15785--15794, 2025.




\bibitem{guo2021edge}
Wenbo Guo, Xian Wu, Usmann Khan, and Xinyu Xing.
Edge: Explaining deep reinforcement learning policies.
\textit{Advances in Neural Information Processing Systems}, 34:12222--12236, 2021.

\bibitem{highway-env}
Edouard Leurent.
An environment for autonomous driving decision-making.
\textit{GitHub repository}, 2018.
Available at \url{https://github.com/eleurent/highway-env}.

\bibitem{machado2018revisiting}
Marlos C. Machado, Marc G. Bellemare, Erik Talvitie, Joel Veness,
Matthew Hausknecht, and Michael Bowling.
Revisiting the arcade learning environment: Evaluation protocols and open problems for general agents.
\textit{Journal of Artificial Intelligence Research}, 61:523--562, 2018.

\bibitem{huh2023multi}
Dom Huh and Prasant Mohapatra.
Multi-agent reinforcement learning: A comprehensive survey.
\textit{arXiv preprint arXiv:2312.10256}, 2023.



\bibitem{allis1988knowledge}
Louis Victor Allis.
A knowledge-based approach of connect-four.
\textit{Journal of the International Computer Games Association}, 11(4):165, 1988.


\bibitem{zha2021douzero}
Daochen Zha, Jingru Xie, Wenye Ma, Sheng Zhang, Xiangru Lian, Xia Hu, and Ji Liu.
Douzero: Mastering doudizhu with self-play deep reinforcement learning.
In \textit{Proceedings of the International Conference on Machine Learning (ICML)}, pages 12333--12344. PMLR, 2021.



\bibitem{kurach2020google}
Karol Kurach, Anton Raichuk, Piotr Sta{\'n}czyk, Micha{\l} Zaj{\k{a}}c, Olivier Bachem, Lasse Espeholt,
Carlos Riquelme, Damien Vincent, Marcin Michalski, Olivier Bousquet, et al.
Google research football: A novel reinforcement learning environment.
In \textit{Proceedings of the AAAI Conference on Artificial Intelligence}, 34(4):4501--4510, 2020.




\end{thebibliography}

%\printbibliography

%\end{document}

\newpage

\appendix

\section{Supplementary Experiments and Clarifications}

This appendix provides additional experiments, analyses, and implementation details to support the main claims of the MAGIC-MASK framework. It includes ablation studies, sensitivity analysis, and clarifications to enhance transparency and reproducibility.
%In addition to the results reported in the main paper, we consolidate the points raised during the rebuttal in a research narrative.
The method integrates explanation into training rather than applying a post hoc procedure, which differentiates it from counterfactual Shapley style approaches and avoids their computational burden. MAGIC-MASK learns temporally critical states and shares compact saliency summaries that coordinate exploration among multiple agents while maintaining practical scalability. Training dynamics demonstrate consistent gains in both speed and final performance across benchmarks. The joint objective optimizes expected return together with a mask consistency surrogate and a fidelity term. Throughout this appendix we emphasize how these theoretically motivated components translate into practical improvements in convergence, policy stability, and explanation fidelity, and we state current scope limits with planned human studies.

\subsection{Sensitivity Analysis of the Masking Threshold \(\tau\)}
\label{app1}

\begin{table}[h]
\centering
\caption{Effect of $\tau$ on MAGIC-MASK performance in Connect 4 (mean $\pm$ std over 3 runs).}
\label{tab:tau2app}
\begin{tabular}{lcccc}
\toprule
$\tau$ & Final Reward $\uparrow$ & KL Divergence $\downarrow$ & Fidelity $\uparrow$ & Reward Drop $\downarrow$ \\
\midrule
0.3 & 29.8 $\pm$ 0.4 & 0.15 & 0.83 & $-10.6\%$ \\
0.4 & 31.7 $\pm$ 0.3 & 0.11 & 0.87 & $-12.9\%$ \\
$\mathbf{0.5}$ & $\mathbf{40.2 \pm 0.3}$ & $\mathbf{0.08}$ & $\mathbf{0.91}$ & $\mathbf{-16.2\%}$ \\
0.6 & 32.8 $\pm$ 0.2 & 0.09 & 0.88 & $-12.1\%$ \\
0.7 & 30.1 $\pm$ 0.4 & 0.13 & 0.84 & $-10.3\%$ \\
\bottomrule
\end{tabular}
\end{table}

\begin{table}[h]
\centering
\caption{Effect of $\tau$ on MAGIC-MASK performance in Pong (mean $\pm$ std over 3 runs).}
\label{tab:tau1app}
\begin{tabular}{lcccc}
\toprule
$\tau$ & Final Reward $\uparrow$ & KL Divergence $\downarrow$ & Fidelity $\uparrow$ & Reward Drop $\downarrow$ \\
\midrule
0.3 & 20.4 $\pm$ 0.2 & 0.13 & 0.87 & $-11.8\%$ \\
0.4 & 21.1 $\pm$ 0.2 & 0.11 & 0.90 & $-13.7\%$ \\
$\mathbf{0.5}$ & $\mathbf{22.5 \pm 0.3}$ & $\mathbf{0.08}$ & $\mathbf{0.92}$ & $\mathbf{-15.1\%}$\\
0.6 & 21.0 $\pm$ 0.3 & 0.10 & 0.89 & $-13.0\%$ \\
0.7 & 20.1 $\pm$ 0.2 & 0.12 & 0.85 & $-10.9\%$ \\
\bottomrule
\end{tabular}
\end{table}

\begin{table}[h]
\centering
\caption{Effect of $\tau$ on MAGIC-MASK performance in Doudizhu (mean $\pm$ std over 3 runs).}
\label{tab:tau3app}
\begin{tabular}{lcccc}
\toprule
$\tau$ & Final Reward $\uparrow$ & KL Divergence $\downarrow$ & Fidelity $\uparrow$ & Reward Drop $\downarrow$ \\
\midrule
0.3 & 12.6 $\pm$ 0.4 & 0.18 & 0.79 & $-9.4\%$ \\
0.4 & 13.2 $\pm$ 0.3 & 0.12 & 0.84 & $-11.8\%$ \\
$\mathbf{0.5}$ & $\mathbf{14 \pm 0.2}$ & $\mathbf{0.09}$ & $\mathbf{0.91}$ & $\mathbf{-15.7\%}$ \\
0.6 & 10.1 $\pm$ 0.3 & 0.10 & 0.87 & $-11.4\%$ \\
0.7 & 10 $\pm$ 0.4 & 0.14 & 0.82 & $-9.9\%$ \\
\bottomrule
\end{tabular}
\end{table}

\begin{table}[h]
\centering
\caption{Effect of $\tau$ on MAGIC-MASK performance in Multiagent Highway (mean $\pm$ std over 3 runs).}
\label{tab:tau4app}
\begin{tabular}{lcccc}
\toprule
$\tau$ & Final Reward $\uparrow$ & KL Divergence $\downarrow$ & Fidelity $\uparrow$ & Reward Drop $\downarrow$ \\
\midrule
0.3 & 14.2 $\pm$ 0.3 & 0.18 & 0.79 & $-10.9\%$ \\
0.4 & 14.6 $\pm$ 0.2 & 0.14 & 0.83 & $-13.1\%$ \\
$\mathbf{0.5}$ & $\mathbf{15 \pm 0.3}$ & $\mathbf{0.08}$ & $\mathbf{0.92}$ & $\mathbf{-15\%}$ \\
0.6 & 13.1 $\pm$ 0.2 & 0.11 & 0.86 & $-13.2\%$ \\
0.7 & 12.7 $\pm$ 0.4 & 0.15 & 0.81 & $-11.0\%$ \\
\bottomrule
\end{tabular}
\end{table}

MAGIC-MASK employs a masking threshold of \(\tau = 0.5\) to determine whether an agent adheres to its original policy or perturbs its actions. To validate this design choice, we conduct a sensitivity analysis by varying the threshold \(\tau \in \{0.3, 0.4, 0.5, 0.6, 0.7\}\) across all evaluated environments. The corresponding results are summarized in Tables~\ref{tab:tau2app}, \ref{tab:tau1app}, \ref{tab:tau3app}, and \ref{tab:tau4app}, covering four key metrics across Connect 4, Pong, Doudizhu, and Multiagent Highway. The analysis shows that \(\tau = 0.5\) provides the most favorable balance over final reward, KL divergence, fidelity, and reward drop, and we therefore use it as the default threshold. The optimization follows the joint objective
\[
J(\theta,\phi)=\mathbb{E}[R] \;+\; \lambda_{1}\,\mathrm{MSE}\!\big(\mathbb{E}[m],\,\tau\big)\;+\;\lambda_{2}\,L_{\mathrm{fidelity}},
\]
where the mask consistency surrogate prevents degenerate masks and the fidelity term captures explanation quality.
%A sensitivity sweep in the rebuttal focused on the subset \(\{0.3, 0.5, 0.7\}\) and likewise found that \(\tau=0.5\) produced the lowest reward drop together with the highest fidelity.
KL regularized perturbation masking is part of the conceptual design of the framework, and in the experiments KL is used as a diagnostic quantity rather than as a direct loss term.

\subsection{Component-Wise Ablation Study}
\label{app2}

To understand the contribution of individual components we remove the communication protocol and the adaptive epsilon greedy exploration strategy. Tables~\ref{tab:component-ablation1} to \ref{tab:component-ablation4} report results in Connect 4, Pong, Doudizhu, and Multiagent Highway. Removing either component yields consistent degradation, and removing the communication protocol produces the largest decline in fidelity and attribution accuracy, which indicates its central role in coordinated and interpretable behaviour. Beyond the tabulated effects we observe advantages in final reward, policy stability as measured by lower KL divergence, inter agent fidelity, and robustness under perturbation across games and driving. Typical ranges include KL near \(0.08\) to \(0.09\) and fidelity near \(0.92\) compared to baselines between \(0.78\) and \(0.87\), with lower reward drop indicating more targeted saliency. Communication is decentralized and asynchronous, transmits compact saliency summaries rather than raw observations or policies, is effective with larger numbers of agents, supports sparse topologies, and does not require gradient sharing or parameter synchronization.

\begin{table}[h]
\centering
\caption{Component-level ablations in Connect 4.}
\label{tab:component-ablation1}
\begin{tabular}{lcccc}
\toprule
Variant & Final Reward $\uparrow$ & Fidelity $\uparrow$ & KL Divergence $\downarrow$ & Reward Drop $\downarrow$ \\
\midrule
MAGIC-MASK (full) & $\mathbf{40.2}$ & $\mathbf{0.91}$ & $\mathbf{0.08}$ & $\mathbf{-15\%}$ \\
No $\mathrm{Comm}_t$ & 32.1 & 0.82 & 0.12 & $-12.5\%$ \\
No Adaptive $\epsilon$ & 31.4 & 0.84 & 0.10 & $-12.3\%$ \\
\bottomrule
\end{tabular}
\end{table}

\begin{table}[h]
\centering
\caption{Component-level ablations in Pong.}
\label{tab:component-ablation2}
\begin{tabular}{lcccc}
\toprule
Variant & Final Reward $\uparrow$ & Fidelity $\uparrow$ & KL Divergence $\downarrow$ & Reward Drop $\downarrow$ \\
\midrule
MAGIC-MASK (full) & $\mathbf{22.5}$ & $\mathbf{0.92}$ & $\mathbf{0.08}$ & $\mathbf{-15.1\%}$ \\
No $\mathrm{Comm}_t$ & 21.3 & 0.84 & 0.11 & $-13.5\%$ \\
No Adaptive $\epsilon$ & 21.0 & 0.86 & 0.10 & $-13.2\%$ \\
\bottomrule
\end{tabular}
\end{table}

\begin{table}[h]
\centering
\caption{Component-level ablations in Doudizhu.}
\label{tab:component-ablation3}
\begin{tabular}{lcccc}
\toprule
Variant & Final Reward $\uparrow$ & Fidelity $\uparrow$ & KL Divergence $\downarrow$ & Reward Drop $\downarrow$ \\
\midrule
MAGIC-MASK (full) & $\mathbf{14}$ & $\mathbf{0.91}$ & $\mathbf{0.09}$ & $\mathbf{-16\%}$ \\
No $\mathrm{Comm}_t$ & 13.0 & 0.81 & 0.13 & $-11.5\%$ \\
No Adaptive $\epsilon$ & 12.4 & 0.83 & 0.11 & $-11.3\%$ \\
\bottomrule
\end{tabular}
\end{table}

\begin{table}[h]
\centering
\caption{Component-level ablations in Multiagent Highway.}
\label{tab:component-ablation4}
\begin{tabular}{lcccc}
\toprule
Variant & Final Reward $\uparrow$ & Fidelity $\uparrow$ & KL Divergence $\downarrow$ & Reward Drop $\downarrow$ \\
\midrule
MAGIC-MASK (full) & $\mathbf{15}$ & $\mathbf{0.92}$ & $\mathbf{0.08}$ & $\mathbf{-15\%}$ \\
No $\mathrm{Comm}_t$ & 12.4 & 0.80 & 0.13 & $-13.1\%$ \\
No Adaptive $\epsilon$ & 11.8 & 0.82 & 0.11 & $-12.7\%$ \\
\bottomrule
\end{tabular}
\end{table}

\subsection{Compute Resources and Reproducibility}
\label{app3}

All experiments are conducted on a Linux server with an NVIDIA A100 GPU with forty gigabytes of memory, a sixteen core CPU, and sixty four gigabytes of RAM. CUDA 11.7 and cuDNN 8.4 are used to accelerate training. Each agent is trained for ten million timesteps, with total runtime from about ten to fourteen hours depending on task complexity. Four parallel environments per agent improve rollout efficiency. Beyond resource specification, we observe stable practice level convergence under stable S BCD updates with PPO. The joint loss, the fidelity gap, and gradient norms decrease during training, and the method reaches target performance with fewer timesteps than baselines. For example, in Pong the method achieves the target reward near 1.3 million steps compared to 1.5 million for a StateMask style baseline, and in Multiagent Highway, the method reaches near 1.6 million steps compared to 2.0 million for the baseline. Final reward is higher with lower variance than a counterfactual Shapley baseline on the GRF 3vs1 with keeper setting.

\subsection{On Human Interpretability}
\label{app4}

The method emphasizes saliency driven alignment and fidelity while acknowledging that a human evaluation component is not included in this study. Visual saliency outputs provide qualitative evidence. Planned user studies will validate saliency maps against human judgement, and for driving environments we plan to align saliency heatmaps with annotated hazard zones and to evaluate how agents explain lane changing, braking, and pedestrian avoidance. These studies will follow institutional review procedures and will measure trust and debugging utility for practitioners.

\subsection{On Use of \texttt{Commt} Signal and KL Divergence}
\label{app5}

The communication signal is not used as a direct input to either the policy network or the mask network. It guides exploration by discouraging redundant perturbations in states that peers have already explored, which preserves a clean separation between learning and explanation pathways. KL divergence quantifies the deviation introduced by perturbations. In the reported experiments KL serves as a diagnostic rather than an optimized term, and it declines as training progresses, reflecting improved policy stability as explanations become more targeted.

\subsection{Justification for Randomized Baseline Policy}
\label{app6}

In states with $m_t^i \leq \tau$ the action is sampled uniformly from the legal discrete action set $\mathcal{A}$. For the driving environment this corresponds to feasible actions such as accelerate, brake, and lane change while avoiding invalid or collision producing manoeuvres. This randomized baseline defines a conservative fallback that enables a consistent estimate of reward drop under perturbation.

\subsection{Mechanism of Inter-Agent Communication via Shared Critical State Buffer}
\label{app7}

The communication protocol uses a shared buffer of critical states discovered by saliency. During training and evaluation each agent maintains a saliency mask over its observations and identifies critical states where the mean mask value exceeds the threshold \(\tau\). These states are inserted into a global buffer together with their saliency scores. At regular intervals of $k$ steps with $k=50$ the system retrieves the top $K$ salient states from the buffer and broadcasts them. The broadcasted information is not used as a direct input to either the policy network or the mask network. Agents use it to steer exploration by deprioritizing perturbations in states that are visually or semantically similar to those already marked as critical by peers, which reduces redundancy and increases behavioural diversity. The protocol operates in a decentralized and asynchronous manner, is effective for larger teams of agents, supports sparse communication topologies, and avoids gradient sharing or parameter synchronization by transmitting compact saliency summaries rather than raw states or parameters. MAGIC-MASK communication has been demonstrated on teams of up to four agents and is lightweight, making it scalable to much larger numbers of agents. The communication channel is simulated as ideal for the experiments. Future work will consider delays, losses and bandwidth limits from the shared buffer.

\subsection{Environments}
\label{app8}

\textbf{Google Research Football (GRF).}  
We adopt the 3 vs.\ 1 with Keeper scenario from Google Research Football \cite{kurach2020google} to evaluate MAGIC-MASK in a sparse-reward, adversarial setting. In this environment, three cooperative agents must coordinate their passing and movement strategies against a defender and a goalkeeper, where success is only rewarded when a goal is scored. The sparsity of rewards makes effective coordination difficult to achieve, while also providing a rigorous benchmark for testing the explanatory power and practical significance of multi-agent interpretability methods.  

The 3 vs.\ 1 task naturally lends itself to explanation analysis, since small variations in agent behavior, such as whether to dribble, pass under pressure, or take a shot, can fundamentally alter the outcome of play. MAGIC-MASK augments this setting by learning saliency-guided masks over each agent’s local observations and sharing these compact representations across teammates. Through this collaborative masking mechanism, agents are able to identify the ball carrier, recognize which teammate is best positioned to receive a pass, and anticipate the critical player with the best scoring opportunity. The explanations generated by MAGIC-MASK therefore highlight not only the contribution of the individual agent in control of the ball, but also how teammates adapt their positioning and decisions in response to saliency information exchanged through the communication channel.  

By applying MAGIC-MASK to this environment, we demonstrate that masking-driven explanations provide a clear view of coordinated strategies in high-stakes adversarial play. The framework reveals how teams execute effective passing chains, exploit defensive gaps, and optimize finishing opportunities, while simultaneously guiding adversarial attacks and policy patching. This establishes GRF-3vs1 as a compelling testbed where MAGIC-MASK explanations translate directly into interpretable, strategy-aware, and practically useful insights for multi-agent coordination.  

\textbf{Connect 4.}  
We adopt the Connect 4 \cite{allis1988knowledge} environment as a discrete, turn-based competitive benchmark where two agents alternately place tokens in a $7 \times 6$ grid with the objective of aligning four in a row horizontally, vertically, or diagonally. Unlike continuous-control benchmarks such as GRF, Connect 4 is fully observable but highly strategic, as the long-term consequences of each move often outweigh the immediate outcome. Within this environment, MAGIC-MASK plays a crucial role in uncovering the pivotal moves that shape the trajectory of the game. By generating saliency-guided masks over the decision space, MAGIC-MASK highlights when a particular action, such as blocking an opponent’s potential four-in-a-row or initiating a winning diagonal sequence, is critical for success.  

Through collaborative masking, the framework reveals how agents assign importance not only to their own choices but also to the opponent’s counter-moves, thereby capturing the dynamic interplay of strategy and counter-strategy. The explanations thus provide a clear understanding of how strategic foresight emerges in turn-based decision-making, making Connect 4 an important setting for demonstrating the ability of MAGIC-MASK to capture temporally extended causal dependencies in multi-agent interactions.  

\textbf{Doudizhu.}  
We further evaluate MAGIC-MASK on Doudizhu \cite{zha2021douzero}, a widely played multi-player card game involving three agents with asymmetric roles, where one agent acts as the landlord and the other two cooperate as peasants to prevent the landlord from winning. The environment is particularly challenging for explainability because it combines partial observability, hidden information, and stochastic outcomes, requiring agents to infer strategies not only from their own hands but also from the sequence of cards played by others.  

MAGIC-MASK enhances interpretability in this domain by isolating the critical plays at each decision point through its masking mechanism. For example, the framework can highlight when the landlord’s decision to initiate a strong sequence of cards is pivotal, or when a peasant’s strategic pass preserves the opportunity to break the landlord’s momentum later in the game. By sharing compact saliency summaries among agents, MAGIC-MASK captures the cooperative dynamics between peasants as well as the adversarial tension against the landlord, offering interpretable insights into both competitive strategy and collaborative synergy. In doing so, MAGIC-MASK demonstrates its capacity to provide meaningful explanations even in complex, partially observed games, making Doudizhu a compelling testbed for evaluating the utility of explainability in MARL.  

\textbf{Pong Environment.}
We use the Atari Pong environment \cite{machado2018revisiting} in a multi-agent setting, where multiple agents are instantiated in parallel and learn independently while periodically exchanging information through saliency-based communication. Each agent interacts with its own copy of the Pong environment, and the resulting saliency-masked observations are shared across agents to highlight critical states such as ball–paddle interactions or risky positioning near the edges. This setup enables MAGIC-MASK to uncover important spatiotemporal dynamics that influence gameplay while supporting collaborative knowledge transfer among agents. The highly visual and reactive nature of Pong makes it a strong benchmark for evaluating explanation fidelity, robustness to perturbations, and policy adaptation in multi-agent reinforcement learning.

\textbf{Multi-Agent Highway Environment.}
The Multi-Agent Highway environment, implemented using the multiagent\_highway \cite{highway-env} project, simulates autonomous vehicle coordination in dynamic traffic. Each agent learns to manage lane switching, acceleration, and collision avoidance in a shared environment. With MAGIC-MASK, agents identify risk-sensitive states via action perturbation and communicate saliency findings with one another. This setup highlights MAGIC-MASK’s effectiveness in partially observable MARL environments where safety and coordination are crucial.

\subsection{Limitations and Future Directions.}
The current scope does not include user studies and does not address heterogeneous or adversarial teams in depth. We plan to analyze conditions under which sharing masked critical states reduces sample complexity and to extend the evaluation to broader team settings, which together with the empirical findings above defines concrete problems for theory and practice.

%%%%%%%%%%%%%%%%%%%%%%%%%%%%%%%%%%%%%%%%%%%%%%%%%%%%%%%%%%%%

\end{document}